\documentclass[conference]{IEEEtran}
\IEEEoverridecommandlockouts

\usepackage{cite}
\usepackage{amsmath,amssymb,amsfonts}
\usepackage{algorithmic}
\usepackage{graphicx}
\usepackage{textcomp}
\usepackage{xcolor}
\usepackage{array}
\usepackage{makecell}
\def\BibTeX{{\rm B\kern-.05em{\sc i\kern-.025em b}\kern-.08em
    T\kern-.1667em\lower.7ex\hbox{E}\kern-.125emX}}
\begin{document}

\title{Unmasking Digital Falsehoods: A Comparative Analysis of LLM-Based Misinformation Detection Strategies
	\thanks{$^\ast$ Corresponding author: tianyihuang@berkeley.edu}
}

\author{
\IEEEauthorblockN{Tianyi Huang\textsuperscript{$\ast$}}
\IEEEauthorblockA{\textit{Department of Electrical Engineering and Computer Sciences} \\
	\textit{University of California, Berkeley}\\
	Berkeley, CA, USA \\
	tianyihuang@berkeley.edu}
\\    
\IEEEauthorblockN{Peiyang Yu}
\IEEEauthorblockA{\textit{The Information Networking Institute} \\
\textit{Carnegie Mellon University}\\
Pittsburgh, PA, USA \\
peiyangy@alumni.cmu.edu } 
\and
\IEEEauthorblockN{Jingyuan Yi}
\IEEEauthorblockA{\textit{The Information Networking Institute} \\
\textit{Carnegie Mellon University}\\
Pittsburgh, PA, USA \\
jingyuay@alumni.cmu.edu}
\\
\IEEEauthorblockN{Xiaochuan Xu}
\IEEEauthorblockA{\textit{The Information Networking Institute} \\
	\textit{Carnegie Mellon University}\\
	Pittsburgh, PA, USA \\
	xiaochux@alumni.cmu.edu}
}

\maketitle

\begin{abstract}
The proliferation of misinformation on social media has raised significant societal concerns, necessitating robust detection mechanisms. Large Language Models such as GPT-4 and LLaMA2 have been envisioned as possible tools for detecting misinformation based on their advanced natural language understanding and reasoning capabilities. This paper conducts a comparison of LLM-based approaches to detecting misinformation between text-based, multimodal, and agentic approaches. We evaluate the effectiveness of fine-tuned models, zero-shot learning, and systematic fact-checking mechanisms in detecting misinformation across different topic domains like public health, politics, and finance. We also discuss scalability, generalizability, and explainability of the models and recognize key challenges such as hallucination, adversarial attacks on misinformation, and computational resources. Our findings point towards the importance of hybrid approaches that pair structured verification protocols with adaptive learning techniques to enhance detection accuracy and explainability. The paper closes by suggesting potential avenues of future work, including real-time tracking of misinformation, federated learning, and cross-platform detection models.
\end{abstract}


\section{Introduction}

\subsection{Background and Motivation}

1) The Rise of Misinformation on Social Media
The Social media's proliferation has transformed information dissemination by enabling instant access but also accelerated fake news, rumors, and conspiracy theories~\cite{1}. Large language models (LLMs) exacerbate this by mass-producing convincing misinformation~\cite{2}.

Misinformation’s societal impacts include eroded public trust, disrupted democratic processes (e.g., 2016 U.S. election polarization), and public health crises like COVID-19 vaccine hesitancy~\cite{3}. It also threatens economic stability and scientific progress by fostering distrust in markets and research~\cite{4}.

Multimodal misinformation (images, videos, audio) further complicates detection, necessitating advanced cross-modal analysis tools~\cite{5}.

2) Role of LLMs in Misinformation Detection

Large language models (LLMs) like GPT-4 and LLaMA show promise in combating misinformation through advanced language understanding and reasoning. They can fact-check claims by cross-referencing factual databases and detecting inconsistencies in text or multimodal content. Ensemble NLP models may further improve detection by refining linguistic coherence and analytical rigor~\cite{6}.
Various techniques have been tried for applying LLMs in identifying misinformation, including:
\begin{itemize}
	\item Prompt Prompt-based Detection: Zero-shot, few-shot, or chain-of-thought prompting to evaluate claim validity.
	\item Agentic Verification: Systems like FactAgent mimic human fact-checking by breaking verification into subtasks~\cite{7}.
	\item Multimodal Analysis: Combining LLMs with vision-language models (LVLMs) to detect inconsistencies in text-image pairs.
\end{itemize}

LLMs face limitations like hallucination risks and reliance on high-quality training data~\cite{8}. Future work must prioritize model transparency, explainability, and adversarial robustness.

\subsection{Research Objectives}

1) Analyzing the Effectiveness of LLMs in Misinformation Detection

The research in this paper is to access the usability of LLMs to detect misinformation in different domains of study, e.g., politics, public health, and finance.

2) Comparing Text-Based vs. Multimodal Approaches

The study tries to compare text-based and multimodal approaches in identifying misinformation based on effectiveness. While text-based models deal with linguistic signals and fact-checking variety alone, multimodal models process textual, visual, and auditory inputs for higher accuracy~\cite{9}.

3) Evaluating Performance Across Datasets and Benchmarks

To give a general idea of the efficacy of LLM, the research will compare different models with benchmarked datasets that are standardized, i.e., FakeNewsNet, SciNews, and MM-COVID. Recent studies on COVID-19 misinformation emphasize the need for datasets that unify multiple modalities, as combining text, images, and user behavior data improves misinformation detection accuracy~\cite{22}.

\subsection{Contributions of This Review}

1) Comprehensive Overview of Existing Detection Approaches

The paper offers a unified overview of state-of-the-art detection algorithms of misinformation, ranging from text to multimodal and hybrid.

1) Performance-Based Comparison of Leading Models

Comparative analysis shall be conducted amongst leading LLMs for detection of misinformation as per key performance parameters such as accuracy, precision, recall, and explainability.

2) Identification of Key Challenges and Future Directions

Identifying such major challenges for detecting misinformation as model prediction bias, evolving misinformation methods, and scalability, the review would recommend future research directions on how LLM-based detection mechanisms can be optimized and made adaptive.

\section{Taxonomy of Misinformation and Detection Challenges}

\subsection{Types of Misinformation}

\begin{table*}[!ht]
	\centering
	\caption{} \label{tab:1}
	\begin{tabular}{m{2cm}<{\centering}m{5cm}<{\centering}m{5cm}<{\centering}}
		\hline
		Category	&	Definition	&	Characteristics	\\
		\hline
		Fake News	&	Deliberately fabricated or misleading information presented as legitimate news to deceive audiences.	&	Exploits sensationalism and emotional appeal to drive engagement and influence public opinion.	\\
		Rumors	&	Unverified information that spreads rapidly.	&	Fueled by uncertainty and speculation, it often lacks credible sources.	\\
		Conspiracy Theories	&	Propose hidden, malevolent forces behind significant events.	&	Lack credible evidence but resonate with those seeking explanations.	\\
		Disinformation	&	Intentionally spread falsehoods aimed at misleading or manipulating audiences.	&	Often politically or ideologically motivated, strategically crafted to deceive.	\\
		Misinformation	&	Unintentionally shared incorrect information without malicious intent.	&	Can still contribute to misinformation spread and undermine trust in reliable sources.	\\
		Clickbait	&	Sensationalized headlines designed to attract engagement.	&	Prioritizing clicks and views over factual accuracy, may mislead audiences.	\\
		Propaganda	&	Information crafted to influence public perception and promote agendas.	&	Used for political, ideological, or persuasive purposes, often one-sided or biased.	\\
		\hline
	\end{tabular}
\end{table*}

\subsection{Characteristics of Misinformation}

1) Spread Patterns (Viral Nature, Echo Chambers)
Misinformation spreads rapidly due to social media algorithms that boost engaging material, which leads to viral spreading.

2) Psychological Influence and User Biases

Cognitive biases such as confirmation bias and illusory truth effect are responsible for the perpetuation of the acceptance of misinformation~\cite{10}. Users verify information that reinforces their pre-existing beliefs and will find it easier to reject corrections. Emotion plays a key role in misinformation spread. Studies show LLaMA3 excels at identifying emotions in short text, which could aid in detecting emotionally charged misinformation~\cite{11,12}.

3) Multimodal Misinformation (Text, Image, Video)

Modern misinformation typically uses a combination of text, images, and video for greater credibility and effectiveness. The multimodal format makes detection more difficult because sophisticated cross-modal analysis methods are required~\cite{13}. Context is key in geospatial misinformation, as location verification studies show surrounding tweets improve model accuracy~\cite{14}.

\subsection{Challenges in Misinformation Detection}

1) Contextual Ambiguity and Intent Detection

Satire, opinion, and deliberate misinformation are hard to identify due to contextual ambiguity. Differences in culture, linguistic subtleties, and diverse perceptions prevent machines from properly inferring intent.

2) Scalability and Generalizability of Detection Methods

Misinformation keeps on evolving, and therefore detection also has to be adaptive and scalable. Generalization across different languages, domains, and platforms is a key challenge because models trained on specific data may not fare well in different environments. Layer-wise task encoding in LLMs affects misinformation detection, as complex inferential reasoning may require deeper layers for accurate assessment~\cite{15}. Multi-event extraction models like DEEIA improve contextual understanding and efficiency, which could enhance misinformation detection across related events~\cite{20}.

3) Resource Limitations in Real-Time Detection

Real-time misinformation detection requires significant computational resources and infrastructure, which requires high processing power to handle massive volumes of content in different modalities.

\subsection{Comparison of Detection Challenges Across Approaches}
1) Challenges in Text-Based Approaches

Text-based detection mechanisms struggle with sarcasm, irony, and hidden communication, which can hide the real intent of messages. In addition, a consideration of language patterns alone could theoretically fail even to detect contextual subtleties of misinformation.

2) Challenges in Multimodal Approaches

Multimodal systems that include text, image, and video also suffer from integration problems. Heterogeneous modalities of data, resolving inconsistencies, and processing multimedia content transformed demand for sophisticated methods and immense computing resources. Cross-modal integration challenges in misinformation detection parallel those in audio-visual IVD, where single-modality analysis is insufficient~\cite{16}.

3) Limitations in Hybrid Approaches

Hybrid methods combining detection approaches are more precise but more difficult to apply and interpret.

\section{Comparative Analysis of LLM-Based Misinformation Detection Methods}

\subsection{Text-Based Approaches}
\subsubsection{Hybrid Attention Framework: Integrating Statistical and Semantic Features}
The Hybrid Attention Framework for Fake News Detection builds upon LLM-based false information detection by utilizing both textual statistical features and deep semantic comprehension. The model utilizes a hybrid attention mechanism that dynamically attends to appropriate linguistic and structure patterns, improving detection efficiency. Furthermore, attention heat maps and SHAP values provide transparency into model decision-making. While statistical and semantic feature integration enhances detection robustness, feature engineering dependence may limit adaptability to new misinformation trends.

1) FactAgent: Stepwise Fact-Checking with Structured Workflows

FactAgent takes a stepwise, systematic approach to misinformation identification by breaking down fact-checking into tangible sub-tasks. LLMs are leveraged by the model to sequentially check claims against reliable sources of information in an open and transparent fashion.

2) SciNews: Fine-Tuning LLMs for Scientific Misinformation

SciNews is a technology that fights science disinformation through optimizing LLMs on peer-reviewed journal article-based and verifiable source-domain corpora. In doing this, the model more effectively identifies inaccuracy when it comes to health- and science-related content.

3) Prompt Engineering Techniques for Misinformation Detection

Prompt engineering has become an effective misinformation detection methodology in training LLMs to rate text based on more contextual information. Techniques including zero-shot, few-shot, and chain-of-thought prompting enable models to perform fine-grained claim validation with less human intervention.

\subsection{Multimodal Approaches}
1) SNIFFER: Combining Textual and Visual Elements

SNIFFER is a multimodal method that combines textual and visual analysis to detect misinformation on varied social media. The method improves the quality of detection at the cost of feature matching complexity and process complexity.

\begin{figure}
	\centering
	\includegraphics[width=\linewidth]{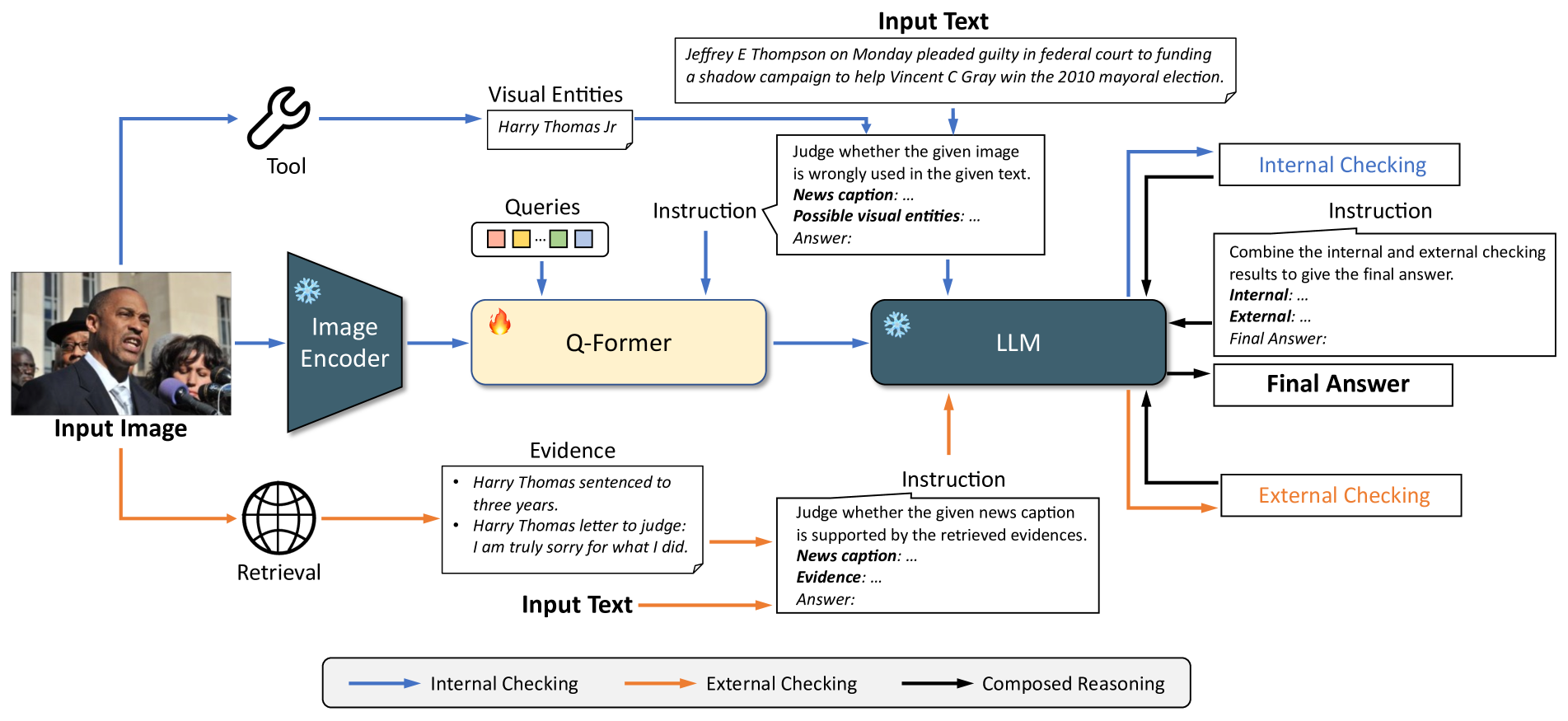}
	\caption{Reference from Qi, Peng, et al. "SNIFFER: Multimodal Large Language Model for Explainable Out-of-Context Misinformation Detection." Proceedings of the IEEE/CVF Conference on Computer Vision and Pattern Recognition. 2024.}
	\label{fig:1}
\end{figure}

2)	LVLM4FV: Image-Text Fusion Models

The LVLM4FV model integrates text- and image-based features to evaluate the credibility of social media content. It has been found to demonstrate promising results in identifying fake media content but requires significant amounts of computational resources and data to deliver best performance.

\begin{figure}
	\centering
	\includegraphics[width=\linewidth]{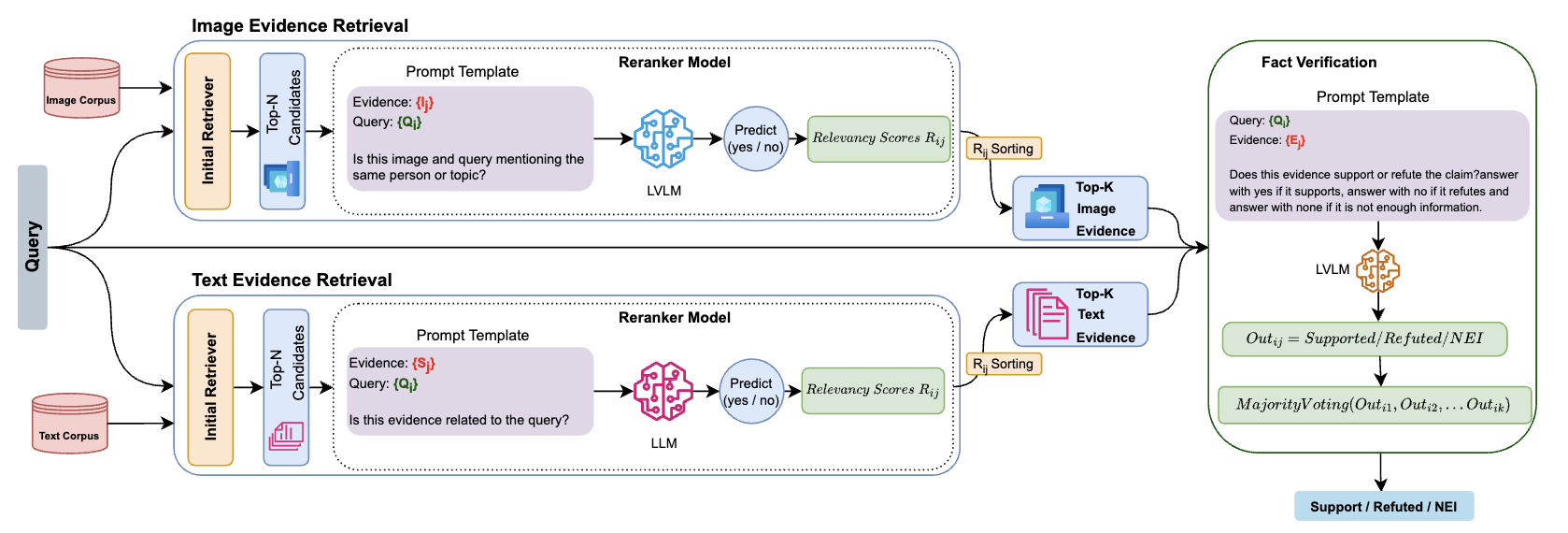}
	\caption{Reference from Tahmasebi, Sahar, Eric M{\"u}ller-Budack, and Ralph Ewerth. "Multimodal misinformation detection using large vision-language models." Proceedings of the 33rd ACM International Conference on Information and Knowledge Management. 2024. }
	\label{fig:2}
\end{figure}

\subsection{Agentic vs. Non-Agentic Models}
1)	Agentic Models (FactAgent, Structured Workflows)

Agentic models such as FactAgent use process workflows to make the process of detecting misinformation easier. Agentic models apply reasoning capabilities and external knowledge bases in a bid to prove claims step by step, making the process explainable but less objective. Agentic models will, however, take longer to respond due to step-by-step processing.

2)	Non-Agentic Models (GPT-4, LLaMa2, Zero-Shot Capabilities)

Non-agentic models such as GPT-4 and LLaMa2 process data in an end-to-end manner without structuring workflows. They are pre-trained and employ prompts for interaction, making them capable of processing content at high speed.

3)	Logical Differences in Processing Workflows

The main difference between agentic and non-agentic models is how they identify misinformation. Agentic models perform a sequence of operations in a defined order in a traceable and reliable manner, while non-agentic models provide immediate feedback by tapping humongous reservoirs of pre-trained knowledge but are not explainable.

\subsection{Comparison of Detection Pipelines}
1)	Zero-Shot vs. Fine-Tuned Approaches

Zero-shot models (GPT-4 leverage pre-trained knowledge for cross-domain adaptability without fine-tuning.

Domain-specific fine-tuned models achieve higher precision in fields like science/medicine but lack generalization, outperforming zero-shot models in targeted tasks.

2)	Performance in Cross-Domain Misinformation Detection

Domain-generalization based misinformation detection is difficult because of the differences in topics, cultures, and languages. Fine-tuning works well on the training set but fails outside.

Zero-shot methods are stronger for different types of misinformation but potentially worse in exploiting domain-specific gimmicks, i.e., fraudulent financial records.

3)	Explainability Differences in Detection Models

Explainability challenges persist: Agentic models (e.g., FactAgent) enhance transparency via structured, source-based verification but sacrifice efficiency, while non-agentic LLMs (e.g., GPT-4) prioritize speed over interpretability. Multimodal models (e.g., SNIFFER) improve accuracy through text-image analysis but struggle with modality alignment. Transparency-efficiency trade-offs persist: agentic models are slower, while black-box LLMs require post hoc explainability tools (e.g., LIME, SHAP). Hybrid approaches integrating fine-tuning, zero-shot learning, and structured reasoning may balance accuracy, flexibility, and transparency across domains.

\begin{table*}[!ht]
	\centering
	\caption{Table Type Styles} \label{tab:2}
	\begin{tabular}{ccccc}
		\hline
		Model Type	&	Example	&	Methodology	&	Strengths	&	Limitation	\\
		\hline
		Pre-train model	&	GPT-4	&	Zero-shot/few-shot learning	&	Generalizability	&	Potential hallucinations	\\
		Agent model	&	FactAgent	&	Structured validation	&	Explainability	&	Limited scalability	\\
		Multimodal model 	&	SNIFFER	&	Multimodal consistency checks	&	Robustness across domains	&	Resource-intensive	\\
		\hline
	\end{tabular}
\end{table*}

\section{Performance Evaluation of LLM-Based Misinformation Detection}
\subsection{Evaluation Metrics}

1)	Standard Metrics

Misinformation detection models are evaluated using accuracy (overall prediction correctness), precision (true misinformation ratio), recall (detection coverage), and F1-score (precision-recall balance). These metrics are critical for assessing text-based and multimodal model effectiveness.

2)	Reliability Measures

To quantify the reliability and agreement of misinformation detection models to some degree, advanced metrics such as Cohen's Kappa and Matthews Correlation Coefficient (MCC) are employed. MCC is a classification performance metric that is overall and particularly effective when dealing with imbalanced data and thus appropriate for misinformation detection.

3)	Model Efficiency

Model efficiency matters when it comes to real-time detection of misinformation. Inference speed and resource consumption (e.g., computation, memory usage) determine feasibility of scalability for LLMs. Techniques such as MG-PTQ utilize graph neural networks to optimize quantization, improving efficiency in low-resource settings without compromising model effectiveness~\cite{17}. Optimization techniques, such as particle swarm optimization (PSO) applied to Transformers, have demonstrated improvements in model efficiency. Similar approaches could be explored to enhance LLM-based misinformation detection~\cite{21}.

\subsection{Benchmark Datasets}

1)	SciNews Dataset

SciNews dataset is tailored to identify scientific misinformation. It includes fact-checked news from high-quality sources and cases of low-quality sources' misinformation, thus making it an ideal dataset to assess the performance of LLMs in scientific domains such as health and climate change misinformation.

2)	FakeNewsNet and LIAR Datasets

FakeNewsNet and LIAR datasets have also been widely adopted in detecting misinformation. FakeNewsNet includes social context information and user interaction data, and LIAR solves the issue of political misinformation and employs fine-grained truth score grading. Both are significant benchmark databases for testing the performance of detection systems.

3)	Multimodal Misinformation Datasets (MM-COVID, PHEME)

The PHEME and MM-COVID data sets have multimodal misinformation, i.e., text, social media metadata, and images. The data sets are important in an attempt to evaluate the performance of multimodal detection algorithms such as SNIFFER and LVLM4FV that process both visual and text content for the purpose of detecting misinformation.

\subsection{Comparative Performance Analysis}

1)	Performance of GPT-4 vs. FactAgent vs. SNIFFER

Comparison of GPT-4, FactAgent, and SNIFFER reveals that while GPT-4 possesses high generalizability and quick response speed, FactAgent enriches structured fact-checking processes that are explainable. SNIFFER is great in multimodal fake news detection but is limited by computational costs due to the need for cross-modal analysis.

\begin{table}[!ht]
	\centering
	\setlength{\tabcolsep}{4pt}
	\caption{Table Type Styles} \label{tab:3}
	\begin{tabular}{ccccc}
		\hline
		Model	&	Dataset	&	Accuracy (\%)	&	F1-Score (\%)	&	Explainability	\\
		\hline
		GPT-4	&	SciNews	&	85.3	&	81.5	&	Low	\\
		FactAgent	&	LIAR	&	91.2	&	87.8	&	High	\\
		SNIFFER	&	MM-COVID	&	88.9	&	85.2	&	Moderate	\\
		\hline
	\end{tabular}
\end{table}

2)	Strengths and Weaknesses Based on Dataset Performance

All the models possess various strengths and weaknesses based on the dataset used. GPT-4 performs well with text-based datasets such as LIAR but not with multimodal disinformation. FactAgent is very suitable for scientific articles, while SNIFFER performs better at image-text contradiction detection but requires enormous computational power.

3)	Explainability vs. Detection Accuracy Trade-Offs

Explainability and detection accuracy are usually at a trade-off in the detection of misinformation. Agentic models such as FactAgent offer more transparency but sacrifice speed and scalability. Non-agentic models such as GPT-4, however, offer fast predictions but are not interpretable, which is a drawback for trust in automated detection systems.

\subsection{Challenges in Performance Evaluation}

1)	Domain-Specific Performance Variability

Misinformation detection models struggle with managing domain-specific variability in performance. A model trained on political misinformation may not work well for the detection of health misinformation since linguistic features and contextualizations are unique.

2)	Bias and Fairness Considerations

Bias and fairness remain primary detection of misinformation challenges. LLMs have a tendency to copy biases in their training data, hence triggering discrimination against certain groups or viewpoints. Bias can be addressed by utilizing different datasets and fairness-aware training procedures to have equitable performance across demographic as well as cultural boundaries.

3)	Handling Evolving Misinformation Trends

Misinformation evolves rapidly, with fresh stories and tactics emerging on a constant basis. Adaptive learning approaches, incremental updates, and human-in-the-loop techniques can enhance model robustness against emerging threats.

\section{Explainability and Interpretability of Detection Models}

\subsection{Importance of Explainability in Misinformation Detection}

1)	User Trust and Decision-Making

Explainability is also essential for developing trust in the models that are utilized to identify misinformation by the users. Transparent and clear explanations enable the users to identify false positives and negatives and hence enable them to reach a reasonable decision.

2)	Transparency in Automated Detection Systems

Transparency is needed to allow auditing and validation of misinformation detection systems. Giving clear, understandable output allows stakeholders to view the procedure whereby conclusions are derived, thereby simplifying detecting bias and limits~\cite{18}. RAG-based methods also contribute to model interpretability by making retrieval paths explicit. RAG-Instruct, for example, enhances instruction diversity and improves structured knowledge integration, which could increase explainability in misinformation detection~\cite{19}.

\subsection{Explainability Techniques}

1)	LIME and SHAP for Model Interpretability

Local Interpretable Model-agnostic Explanations (LIME) and SHapley Additive exPlanations (SHAP) are the most widely applied methods for explanation of misinformation detection models. LIME creates local approximations of model behavior through the utilization of input perturbations and monitoring the effects, and SHAP provides feature importance scores by allocating the contribution of features to push the prediction of the output. They help provide explanations as to why a model predicts content as misinformation.

2)	Integrated Gradients for Deep Models

Integrated Gradients is one of the deep learning models approaches to assigning scores to input features based on their contribution to model prediction. This particularly comes into view for LLM-based misinformation classification, where word-level or sentence-level textual attribution can point to specific words or word sequences toward a misinformation label.

\subsection{Comparison of Explainability Methods Across Models}

1)	Explainability Effectiveness in GPT-4 vs. FactAgent

GPT-4 provides less transparency since it is a black-box model and following how it draws conclusions is difficult. FactAgent, as it applies step-by-step structured verification procedures in an open way, provides greater transparency.

2)	Multimodal Interpretability in SNIFFER

SNIFFER, being an end-to-end text-image model, also possesses some unique challenges to explain itself. SNIFFER requires sophisticated visualization techniques and convergence of modalities to try to give good explanations of multimodal inputs. Despite such challenges, SNIFFER's visual-text alignment modules make it easier to understand compared to text models.

\section{Conclusion}

\subsection{Practical Implications for Misinformation Mitigation}

The findings underscore the need to follow a hybrid approach that leverages both structured and unstructured models to balance accuracy, efficiency, and interpretability. Utilizing real-time monitoring systems, developing multimodal fusion, and making AI-driven misinformation detection systems more transparent are critical to ensuring public trust and efficient content moderation on social media. Collaboration between researchers, policymakers, and platformers is also necessary to deal with new misinformation challenges.

\bibliographystyle{IEEEtran}

\begin{thebibliography}{10}
\providecommand{\url}[1]{#1}
\csname url@samestyle\endcsname
\providecommand{\newblock}{\relax}
\providecommand{\bibinfo}[2]{#2}
\providecommand{\BIBentrySTDinterwordspacing}{\spaceskip=0pt\relax}
\providecommand{\BIBentryALTinterwordstretchfactor}{4}
\providecommand{\BIBentryALTinterwordspacing}{\spaceskip=\fontdimen2\font plus
\BIBentryALTinterwordstretchfactor\fontdimen3\font minus \fontdimen4\font\relax}
\providecommand{\BIBforeignlanguage}[2]{{%
\expandafter\ifx\csname l@#1\endcsname\relax
\typeout{** WARNING: IEEEtran.bst: No hyphenation pattern has been}%
\typeout{** loaded for the language `#1'. Using the pattern for}%
\typeout{** the default language instead.}%
\else
\language=\csname l@#1\endcsname
\fi
#2}}
\providecommand{\BIBdecl}{\relax}
\BIBdecl

\bibitem{1}
B.~Hu, Z.~Mao, and Y.~Zhang, ``An overview of fake news detection: From a new perspective,'' \emph{Fundamental Research}, vol.~5, no.~1, pp. 332--346, 2025.

\bibitem{2}
M.~Chen, L.~Wei, H.~Cao, W.~Zhou, and S.~Hu, ``Explore the potential of llms in misinformation detection: An empirical study,'' in \emph{AAAI 2025 Workshop on Preventing and Detecting LLM Misinformation (PDLM)}.

\bibitem{3}
Y.~Cao, A.~M. Nair, E.~Eyimife, N.~J. Soofi, K.~Subbalakshmi, J.~R. Wullert~II, C.~Basu, and D.~Shallcross, ``Can large language models detect misinformation in scientific news reporting?'' \emph{arXiv preprint arXiv:2402.14268}, 2024.

\bibitem{4}
V.~S. Pendyala and C.~E. Hall, ``Explaining misinformation detection using large language models,'' \emph{Electronics}, vol.~13, no.~9, p. 1673, 2024.

\bibitem{5}
P.~Qi, Z.~Yan, W.~Hsu, and M.~L. Lee, ``Sniffer: Multimodal large language model for explainable out-of-context misinformation detection,'' in \emph{Proceedings of the IEEE/CVF conference on computer vision and pattern recognition}, 2024, pp. 13\,052--13\,062.

\bibitem{6}
X.~Huang, Y.~Wu, D.~Zhang, J.~Hu, and Y.~Long, ``Improving academic skills assessment with nlp and ensemble learning,'' in \emph{2024 IEEE 7th International Conference on Information Systems and Computer Aided Education (ICISCAE)}.\hskip 1em plus 0.5em minus 0.4em\relax IEEE, 2024, pp. 37--41.

\bibitem{7}
X.~Li, Y.~Zhang, and E.~C. Malthouse, ``Large language model agent for fake news detection,'' \emph{arXiv preprint arXiv:2405.01593}, 2024.

\bibitem{8}
C.~Chen and K.~Shu, ``Can llm-generated misinformation be detected?'' \emph{arXiv preprint arXiv:2309.13788}, 2023.

\bibitem{9}
S.~Tahmasebi, E.~M{\"u}ller-Budack, and R.~Ewerth, ``Multimodal misinformation detection using large vision-language models,'' in \emph{Proceedings of the 33rd ACM International Conference on Information and Knowledge Management}, 2024, pp. 2189--2199.

\bibitem{22}
H.~Guo, T.~Huang, H.~Huang, M.~Fan, and G.~Friedland, ``A systematic review of multimodal approaches to online misinformation detection,'' in \emph{2022 IEEE 5th International Conference on Multimedia Information Processing and Retrieval (MIPR)}.\hskip 1em plus 0.5em minus 0.4em\relax IEEE, 2022, pp. 312--317.

\bibitem{10}
Q.~Zeng, G.~Liu, Z.~Xue, D.~Ford, R.~Voigt, L.~Hagen, and L.~Li, ``Sympathy over polarization: A computational discourse analysis of social media posts about the july 2024 trump assassination attempt,'' \emph{arXiv preprint arXiv:2501.09950}, 2025.

\bibitem{11}
J.~Zhang, Z.~Mai, Z.~Xu, and Z.~Xiao, ``Is llama 3 good at identifying emotion? a comprehensive study,'' in \emph{Proceedings of the 2024 7th International Conference on Machine Learning and Machine Intelligence (MLMI)}, 2024, pp. 128--132.

\bibitem{12}
Z.~Mai, J.~Zhang, Z.~Xu, and Z.~Xiao, ``Is llama 3 good at sarcasm detection? a comprehensive study,'' in \emph{Proceedings of the 2024 7th International Conference on Machine Learning and Machine Intelligence (MLMI)}, 2024, pp. 141--145.

\bibitem{13}
J.~Yi, Z.~Xu, T.~Huang, and P.~Yu, ``Challenges and innovations in llm-powered fake news detection: A synthesis of approaches and future directions,'' \emph{arXiv preprint arXiv:2502.00339}, 2025.

\bibitem{14}
Z.~Xiao, Y.~Huang, and E.~Blanco, ``Context helps determine spatial knowledge from tweets,'' in \emph{Findings of the Association for Computational Linguistics: IJCNLP-AACL 2023 (Findings)}, 2023, pp. 149--160.

\bibitem{15}
M.~Jin, Q.~Yu, J.~Huang, Q.~Zeng, Z.~Wang, W.~Hua, H.~Zhao, K.~Mei, Y.~Meng, K.~Ding \emph{et~al.}, ``Exploring concept depth: How large language models acquire knowledge and concept at different layers?'' \emph{arXiv preprint arXiv:2404.07066}, 2024.

\bibitem{20}
W.~Liu, L.~Zhou, D.~Zeng, Y.~Xiao, S.~Cheng, C.~Zhang, G.~Lee, M.~Zhang, and W.~Chen, ``Beyond single-event extraction: Towards efficient document-level multi-event argument extraction,'' \emph{arXiv preprint arXiv:2405.01884}, 2024.

\bibitem{16}
X.~Li, R.~Mohammed, T.~Mangin, S.~Saha, R.~T. Whitaker, K.~E. Kelly, and T.~Tasdizen, ``Joint audio-visual idling vehicle detection with streamlined input dependencies,'' \emph{arXiv preprint arXiv:2410.21170}, 2024.

\bibitem{17}
W.~Liu, Y.~Xiao, D.~Zeng, H.~Zhao, W.~Chen, and M.~Zhang, ``Mixed-precision graph neural quantization for low bit large language models,'' \emph{arXiv preprint arXiv:2501.18154}, 2025.

\bibitem{21}
P.~Yu, J.~Yi, T.~Huang, Z.~Xu, and X.~Xu, ``Optimization of transformer heart disease prediction model based on particle swarm optimization algorithm,'' \emph{arXiv preprint arXiv:2412.02801}, 2024.

\bibitem{18}
T.~Huang, Z.~Xu, P.~Yu, J.~Yi, and X.~Xu, ``A hybrid transformer model for fake news detection: Leveraging bayesian optimization and bidirectional recurrent unit,'' \emph{arXiv preprint arXiv:2502.09097}, 2025.

\bibitem{19}
W.~Liu, J.~Chen, K.~Ji, L.~Zhou, W.~Chen, and B.~Wang, ``Rag-instruct: Boosting llms with diverse retrieval-augmented instructions,'' \emph{arXiv preprint arXiv:2501.00353}, 2024.

\end{thebibliography}

\end{document}